\documentclass{article}

\usepackage{arxiv}

\usepackage[utf8]{inputenc} % allow utf-8 input
\usepackage[T1]{fontenc}    % use 8-bit T1 fonts
\usepackage{hyperref}       % hyperlinks
\usepackage{url}            % simple URL typesetting
\usepackage{booktabs}       % professional-quality tables
\usepackage{amsfonts}       % blackboard math symbols
\usepackage{nicefrac}       % compact symbols for 1/2, etc.
\usepackage{microtype}      % microtypography
\usepackage{lipsum}
\usepackage{graphicx}
\usepackage{xcolor}
\usepackage{multirow}
\usepackage{diagbox}
\usepackage{amsmath}   
\title{Hardware-aware Pruning of DNNs using LFSR-Generated Pseudo-Random Indices}

\author{
  Foroozan.~Karimzadeh \\
  Department of Electrical and Computer Engineering\\
  Georgia Institute of Technology\\
  Atlanta, GA 30332 \\
  \texttt{fkarimzadeh6@gatech.edu} \\
   \And
 Ningyuan.~Cao \\
  Department of Electrical and Computer Engineering\\
  Georgia Institute of Technology\\
  Atlanta, GA 30332 \\
  \texttt{nycao@gatech.edu} \\
   \AND
   Brian.~Crafton \\
  Department of Electrical and Computer Engineering\\
  Georgia Institute of Technology\\
  Atlanta, GA 30332 \\
  \texttt{bcrafton3@gatech.edu} \\
   \And
   Justin.~Romberg\\
  Department of Electrical and Computer Engineering\\
  Georgia Institute of Technology\\
  Atlanta, GA 30332 \\
  \texttt{jrom@ece.gatech.edu} \\
   \And
   Arijit.~Raychowdhury \\
  Department of Electrical and Computer Engineering\\
  Georgia Institute of Technology\\
  Atlanta, GA 30332 \\
  \texttt{arijit.raychowdhury@ece.gatech.edu} \\
}

\begin{document}
\maketitle

\begin{abstract}
Deep neural networks (DNNs) have been emerged as the state-of-the-art algorithms in broad range of applications. To reduce the memory foot-print of DNNs, in particular for embedded  applications,  sparsification  techniques  have  been  proposed.Unfortunately,  these  techniques  come  with  a  large  hardware overhead.  In  this  paper,  we  present  a  hardware-aware  pruning method  where  the  locations  of  non-zero  weights  are  derived  in real-time from a Linear Feedback Shift Registers (LFSRs). Using the  proposed  method,  we  demonstrate  a  total  saving  of  energy and  area  up  to  63.96\%  and  64.23\%  for  VGG-16  network  on down-sampled  ImageNet,  respectively  for  iso-compression-rate and  iso-accuracy.
\end{abstract}

% keywords can be removed

\section{Introduction}
\label{introduction}
Neural networks (NNs) have shown remarkable performance in a broad range of applications, from computer vision \cite{krizhevsky2012imagenet, voulodimos2018deep} to health-care data analysis \cite{stephansen2018neural,miotto2017deep}. These algorithms have gradually evolved to larger models with deeper layers and increasing number of parameters \cite{simonyan2014very}.While the large DNNs are powerful and provides high accuracy for complex tasks, they cannot be easily deployed on embedded mobile applications due to two major problems \cite{zhang2016cambricon}. Firstly, because the models are large and often over-parameterized, it must be stored in an external DRAM. The second major issue is that accessing model parameters from an external DRAM consumes large amounts of energy \cite{han2015deep}. For example, in the 45nm CMOS technology, accessing a 32bit DRAM memory requires 640pJ, which is 3 order of magnitudes higher than a 32bit floating point add operation (0.9 pJ) \cite{han2015deep}. Therefore, it is hard to deploy large DNNs on battery constrained mobile platforms. 

Network compression via pruning techniques is one possible solution to fit large DNNs such as VGG-16 (138M parameters, 520MB) in on-chip SRAM \cite{li2019squeezeflow,parashar2017scnn, lee2008sparse}. However, it is challenging because sparse matrices add extra levels of irregularity to the weights' addresses \cite{li2019squeezeflow,chen2019eyeriss}. In \cite{han2015learning}, a pruning method has been proposed which prunes the connections that have smaller weights than a threshold via an iterative process of pruning and retraining. Although this method prunes the network with no loss of accuracy; the method is heuristic and the threshold values need to carefully selected. Further, resultant matrix lacks structure. In \cite{han2015deep}, the authors prune the network by learning the important connections using a thresholding mechanism and then applying weight sharing and Huffman coding to compress the networks even further. The problem with this method is that they need to store three vectors: (1) the sparse weights matrix's value, (2) the location or address of the non-zero matrix weights and (3) the pointer vector to keep track of the weights in each column. In addition, weight sharing adds another level of indirection and complexity \cite{han2016eie}. In summary, the baseline pruning method is powerful from an algorithm perspective, but its mapping to hardware is inefficient and requires a memory foot-print as high as $2\times$ that of the model size.

In this paper, we present a hardware-aware method to prune dense DNNs which reduces the memory footprint while preserving the original accuracy. We utilize an on-die linear feedback shift register (LFSR) using a known seed, to generate a pseudo random sequence (PRS). Next, we use this PRS to regularize and prune the network. In the last step, we retrain the sparse network so that the model can perform better with the pruned model. In addition, during inference we use the LFSR with the same seed to generate the indices in real-time to perform multiplication between the sparse weight matrix and input/activation vector. Consequently, we no longer need to store the sparse weight addresses -- thereby reducing the memory foot-print significantly.  
 \vspace{-0.1in}
\section{Proposed Hardware-Aware Pruning}
\label{sec:method}

\begin{figure}[!b]
 \vspace{-0.2in}
\centering
  \includegraphics[scale=0.58]{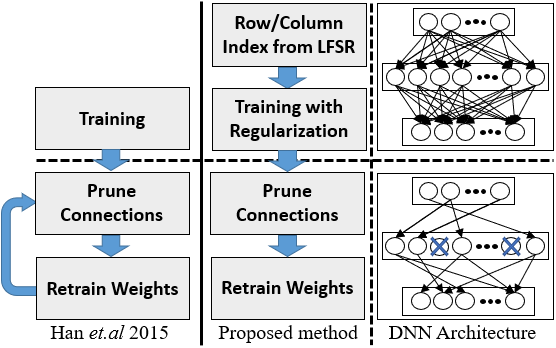}
  \caption{From left to right, training pipeline for baseline (Han\textit{ et.al}, 2015 \cite{han2015learning}), proposed pruning method and an example of pruning the synapses and neurons of a fully connected network.}
  \label{fig:algorithm}

\end{figure}

%%%%%%%%%%%%%%%%%%%%%%%%%%%%%
\begin{figure*}[!t]
\centering
  \includegraphics[scale=0.52]{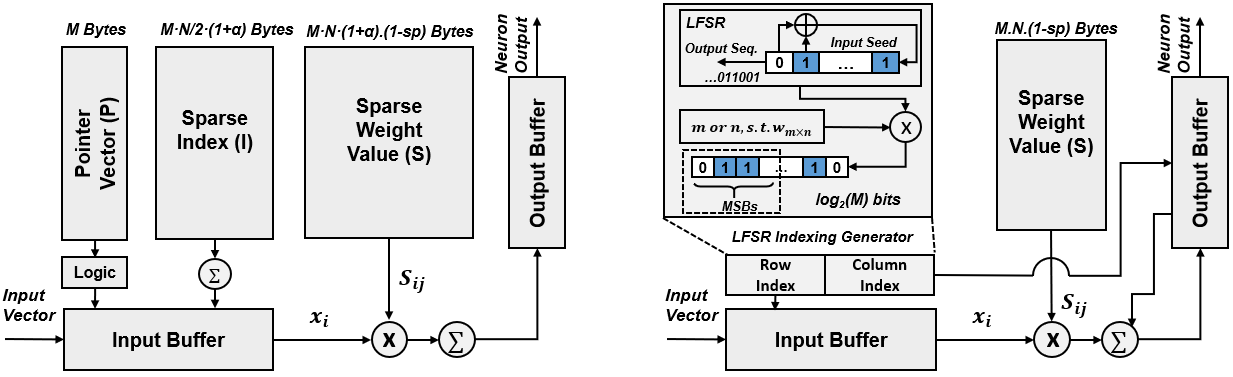}
  \caption{From left to right, the hardware architecture of the baseline and the proposed pruning method, respectively.}
  \label{fig:hardware}
 \vspace{-0.1in}
\end{figure*}
%%%%%%%%%%%%%%%%%%%%%%%%%%%%

Proposed pruning and baseline methods are illustrated in Figure\ref{fig:algorithm}. The proposed pruning method consists four steps.
%each of an iterative approach where each iteration consists of four steps. 
It begins with generating a pseudo-random sequence (PRS) using two LFSRs, one for row indices and another one for column indices. We then use the generated PRS as the indices to sparsify the synapses (i.e. connections) by using standard regularization methods in an iterative approach. The specified synapses are regularized to force them to be zero in the training step and be pruned away in the pruning step. The last step is retraining the sparse model so that the sparse model can provide higher accuracy. Using a PRS for generating the locations of the zero weights in the connectivity matrix provides good accuracy, while making it easier to generate the indices on the fly, instead of being stored in a separate memory sub-bank.

We use Han\textit{ et. al}, 2015 \cite{han2015learning} as the state-of-the-art baseline pruning techniques for the proposed algorithm. In \cite{han2015learning}, illustrated in Figure \ref{fig:algorithm}, a pruning method was proposed to prune the redundant connections in an iterative process based on a threshold. First, the neural network is trained starting with random initial conditions. Next, the connections less than a threshold are pruned iteratively and finally, the pruned network is fine tuned using several epochs of retraining. Interested readers are pointed to \cite{han2015learning} for further details.

\subsection{Linear Feedback Shift Register (LFSR)}

LFSR \cite{mita2002novel} is one of the most commonly used topologies for implementing and generating pseudo random bit sequences~\cite{qasem2018double}. The advantages of using an LFSR to generate the indices are: (1) they can be easily be implemented in hardware, (2) the PRS has key statistical properties that preserves the rank of the generated connectivity matrix \cite{cusick2017cryptographic}. LFSR, consists of an array of flip-flops with an initial state called input seed ($s$), followed by linear feedback performed by several exclusive-or (\textit{XOR}) gates ($c_i$). LFSRs can be mathematically described through $n^{th}$ order characteristic polynomials:
\vspace{-0.05cm}
\begin{equation}
\label{eq1}
x^n + c_{n-1} x^{n-1} + . . . + c_1x + 1 
\end{equation}

% \begin{figure}[tb]
% \centering
%   \includegraphics[scale=0.47]{Fig_lfsr.PNG}
%   \caption{LFSR architecture.}
%   \label{fig:lfsr}
%  \vspace{-0.2in}
% \end{figure}

To obtain the maximum PRS length, the characteristic polynomials have to be primitive \cite{cusick2017cryptographic} where the maximum period without repetition is $2^n-1$.
In the proposed method we utilize two LFSRs, to generate random sequence for the indices of the rows and columns, separately. The row index indicates the address of the element in the input vector while the column address encodes the address of the output vector. 

\subsection{Training with PRS based Regularization}
\label{train}
A fully connected (\textit{FC}) layer of DNN performs the following function:

% \begin{equation}
% \label{eq:weight}   
% Z = W^T x + b 
% \end{equation}
\vspace{-0.3cm}
\begin{equation}
\label{eq:activation}   
a = f (Z)~~~ \text{where} ~~~Z = W^T x + b 
\end{equation}

Where $W$ is a weight matrix, $x$ is an input, $b$ is a vector of bias values, $f$ is a non-linear activation function, typically a Rectified Linear Unit (ReLU) \cite{muralimanohar2009cacti}. For simplification, $b$ can be merged with $W$ by appending an additional column to the end of matrix $W$. Then we can write the above equation as:
\vspace{-0.15cm}
\begin{equation}
\label{eq:fc}   
a = ReLU(\sum_{d=0}^{n-1} W_{cd} x)
\end{equation}

where \textit{c} and \textit{d} are the original weight matrix's indices correspond to rows and columns, respectively.
After selecting the synapses using the PRS sequence, we regularize them during the training process. We apply strong regularization on the selected synapses, based on LFSRs indices, in order to force the network to zero-out these synapses. We have studied both L1 and L2 regularization~\cite{han2015learning} to penalize non-zero weight values. %Using L1 regularization resulting more weights (i.e synapses) to be near zero which gives better performance in terms of accuracy after pruning and before retraining \cite{han2015learning}. While L2 regularization gives the best retraining results. Regularization also prevents over-fitting as it is a random selection of synapses. 
For L2 regularization, a regularizer component is added to the cost (\textit{J}) , as shown in Eq.\ref{eq:regularization} and weights will be updated based on Eq. \ref{eq:weight_update}.
\vspace{-0.2cm}
\begin{equation}
\label{eq:regularization}   
J(W^{[l]}, b^{[l]}) = \frac{1}{m} \sum_{c=1}^{m} L(\hat{y}^{(i)} , y^{(i)}) + \frac{\lambda }{2m} \sum_{l=1}^{L} ||W_{ij}^{[l]}||_F ^2
\end{equation}

\begin{equation}
\label{eq:weight_update}
  W^{[l+1]}=\left\{
  \begin{array}{@{}ll@{}}
    W^{[l]} [1-\frac{\beta \lambda}{m}], & \text{if}\ c, d = i, j \\
    W^{[l]} - \beta dW^{[l]}, & \text{otherwise}
  \end{array}\right.
\end{equation} 

Where \textit{i} and \textit{j} are correspond to the indices from LFSR 1 and LFSR 2 for rows and columns, respectively. $\beta$ is the learning rate. \textit{L} is the layer's number in the network. $\lambda$ is the regularization parameter and can be tuned. Larger $\lambda$ will more penalized the weights values and make them closer to zero.

\subsection{Pruning and Retraining}
\label{prune}

After heavily regularizing the selected weights, a pruning step is employed to make sure that all the selected weights are exactly zero as regularization makes the connection very close to zero, but not exactly zero. With the LFSR based pruning method, the activation computation of Eq. \ref{eq:fc} becomes
\vspace{-0.1in}
\begin{equation}
\label{eq:fc}   
a = ReLU(\sum_{j}^{n-1} S_{ij} x )
\end{equation}

where S is a sparse weight matrix where \textit{i} and \textit{j} belong to the first and second LFSR, respectively. The last step of the process is to retrain the pruned network to fine tune the remaining synapses so that they better compensate for the removed connections.

\subsection{DNN compression}
\label{hardware_implementation}
To fully understand advantages and limitations of baseline and proposed algorithms, we implement both methods in digital hardware (Figure \ref{fig:hardware}). The two circuit diagrams show how hardware resources/operations differ to solve a sparsely connected fully-connected layer, with N input neuron, M output neuron and $Sp$ as sparsity. For the hardware implementation of our proposed method, the LFSR is used to generate the index for the input to be multiplied to the corresponding weight in sparse matrix weight. LFSR generate the pseudo random sequence with values between 1 to $2^N-1$. In order to keep the values between number of input neurons, we multiply the generated value to the length of input neurons and select the most significant bits (MSBs). The goal is to avoid redundant clock cycles when the generated number is greater that the number of neurons.  After doing multiplication and accumulation operations, the result is stored in the  output buffer. The index comes from the LFSR with different input seed which is responsible for generating the sequence for output addresses. The exact number of memory reads from the input and the output buffer depend on the model size as well as the number of multiply and accumulate units available for parallel compute. For the baseline algorithm, the sparse weight matrix is compressed in three vectors including the non-zero values of the weights (S), location of the non-zero weights (I) and a pointer vector to point to the start of each column of the weight matrix (P), that should be saved in the memory. Moreover, each entry bit-width of S and I  is designed to be four-bit or eight-bits, and additional memory usage ratio resulted from limited index representation is denoted by $\alpha$. For instance, if more than 15 zeros appear before a non-zero four-bit entry, a zero is added to vectors S and I.

\section{Results}

\subsection{Simulation results for the proposed pruning algorithm}
The proposed methodology is demonstrated on three pruned networks: LeNet-300-100, LeNet-5 and VGG-16 using MNIST, CIFAR-10 and down-sampled ImageNet \cite{ILSVRC15} data-sets. Training is carried out on Nvidia GTX 1080 Ti  GPUs.  The key parameters for hardware implementation are shown in Table \ref{tab:hardware_param}. Rate of compression along with the top-1 accuracy error before and after pruning for three mentioned models are shown in Table \ref{tab:model_compression}. We observe that the proposed method does not affect the rank of weight matrices (Table \ref{tab:rank}). As the matter of fact the rank in the proposed approach is close to full rank  (as in unpruned models) of the dense matrix. Since the PRS maintain the matrix rank, we infer that the \textit{expressibility} of the weight matrices and accuracy of the network can remain unchanged.

\begin{table}[b]
\centering

\caption{Hardware Parameters.}
\begin{tabular}{c|c}
\label{tab:hardware_param}
     Technology Node & TSMC 65nm \\
     \hline
     Supply Voltage & 1V  \\
     \hline
     Temperature & 25 $^\circ$C \\
     \hline
     Datapath Bitwidth & 8b \\
     \hline
     Index Bitwidth & 4b, 8b   \\
     \hline
     Clock Frequency & 1GHZ   \\
     \hline
     Memory Bank Size & 256B, 512B, 1KB, 4KB   \\
     \hline

\end{tabular}

\end{table}

\begin{table}
\centering

\caption{Number of parameters, pruning and reference accuracy and rate of compression.}
\setlength\tabcolsep{2.5pt}
\begin{tabular}{c|c|c|c}
\label{tab:model_compression}
     Network & Error & Parameters & Compression Rate\\
     \hline
     LeNet-300-100 Unpruned & 4.2\%  & 267K & \\
     \cline{1-3}
     LeNet-300-100 Pruned & 4.9\% & 24K & $11\times$ \\
     \hline
     \hline
     LeNet-5 Unpruned & 1.5\% & 431K &  \\
     \cline{1-3}
     LeNet-5 Pruned & 1.6\% & 43K & $10\times$ \\
     \hline
     \hline
     Modified VGG-16 Unpruned & 48.5\% & 23M & \\
     \cline{1-3}
    Modified VGG-16 Pruned & 52.1\%
    
    % \footnotemark
    & 3.3M & $7\times$\\
     \hline
\end{tabular}
% \footnotetext{Accuracy results for VGG-16 on down-sampled Imagenet is slightly less than the originals because minimal pre-processing and data augmentation were applyed.}

\vspace{-0.2in}
\end{table}

\begin{table}[b]
\centering
\vspace{-0.3in}
\caption{Rank of fully connected layers of LeNet-5 on MNIST in two different sparsity rates for fully connected (FC) layers without pruning and the proposed method.}
\begin{tabular}{c|c|c|c|c|c|c}
\label{tab:rank}
     Network &\multicolumn{3}{c|}{\parbox[c]{1.5cm}{\centering FC layers  (unpruned) }} &\multicolumn{3}{c}{\parbox[c]{1.5cm}{\centering Our Method}} \\
     \cline{1-7}
      {\backslashbox{Sparsity}{Hidden Layer}}& H1& H2& H3 &H1& H2& H3\\
     \hline
     \hline
     
     50(\%) & 120 & 84 & 10 & 118 & 83 & 10 \\
     \hline
     90(\%) & 120 & 84 & 10 & 118 & 82 & 10 \\

\end{tabular}
\vspace{-0.2in}
\end{table}

\subsubsection{ Pruning on Fully Connected Layers}
Large DNNs are over-parameterized; this mostly correspond to the large fully connected layers of these networks. For instance, 124M out of 138M of the parameters are related to the 3 fully connected layer in VGG-16. Because of this, we focused on pruning fully connected layers' connection as they consume most of the energy and memory size from hardware perspective. 

\subsubsection{LeNet on MNIST}
\begin{figure}[!b]

\centering
  \includegraphics[scale=0.5]{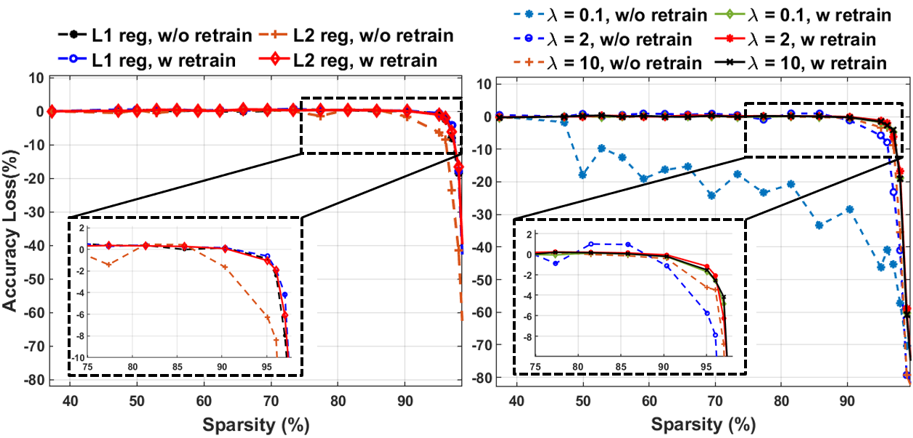}
  \caption{Sparsity patterns for LeNet-300-100 on MNIST. Three different \textit{lambda}, including 0.1, 2 and 10 have been tested and the accuracy loss is shown before and after retraining (right). Trade-off curves for L1 and L2 regularization (left).}
  \label{fig:lfsr_diff_lambda}

\end{figure}

LeNet-300-100 is a fully connected network which has two hidden layers of length 300 and 100 neurons each which achieves 4.9\% error rate on the MNIST dataset. Accuracy loss versus different sparsity rates on MNIST is illustrated in Figure\ref{fig:lfsr_diff_lambda}. Three different $\lambda$ have been tested on the proposed method and the results before and after retraining are illustrated. The results shows that moderate and strong regularization, $\lambda$ equals to 2 and 10, respectively, have better performance before and after retraining. We picked $\lambda$ equals to 2 to trade-off between pruning and convergence of the loss function while preventing over-fitting. We use both L1 and L2 regularization and the results have been illustrated~\ref{fig:lfsr_diff_lambda}. L1 has better performance before retraining while L2 achieves better performance after retraining. 

\begin{figure*}[t!]
\centering
  \includegraphics[scale=0.46]{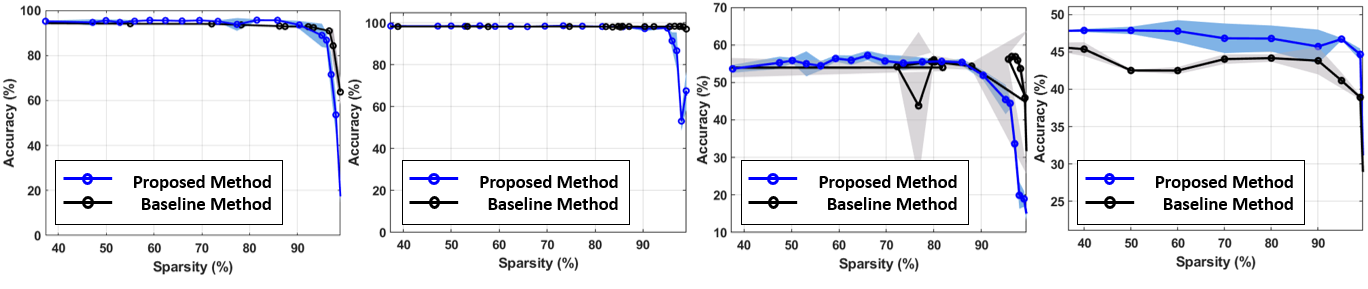}
  \caption{The comparison between the proposed method's $mean \pm std$ accuracy and the baseline (Han \textit{et.al} 2015 \cite{han2015learning}) for different sparsity rates. From left to right figures above are for Lenet-300-100 on MNIST, LeNet-5 on MNIST and LeNet-5 on Cifar10 and VGG-16 on down-sampled ImageNet, respectively. 
  %Based on the results, the proposed method has lower std while preserving the accuracy on more complex dataset (Cifar10).
  }
  \label{fig:lfsr_benchmark}
\end{figure*}

The second model tested on MNIST is a convolutional network, LeNet-5 that has two convolutional layers followed by two fully connected layers. LeNet-5 achieves 1.6\% error rate on MNIST dataset. The accuracy versus different sparsity of LeNet on MNIST is illustrated in Figure \ref{fig:lfsr_benchmark}, which shows that our method achieves the same accuracy as the baseline for different sparsity rates.

\subsubsection{LeNet-5 on CIFAR-10}
 We uses LeNet-5 on CIFAR-10 and the comparison between the $mean \pm std$ accuracy of proposed method and the baseline method for 5 trials is shown in Figure \ref{fig:lfsr_benchmark}. The results shows that the proposed method is more reliable and has less \textit{std} while preserving the original accuracy as it is not based on the thresholding method.   
 
\subsubsection{VGG-16 on down-sampled Imagenet}
We used VGG-16 on ImageNet data \cite{ILSVRC15} with 1000 different classes, but initially down-sampled it to $64\times64$ \cite{oord2016pixel}. Apart from a single crop with no rotation, we have not used any other pre-processing or augmentation. For this dataset we have utilized largest batch size, 32 images/batch, allowed by the GPU memory. We then classified down-sampled ImageNet using VGG-16 with some modification to be fit to the spatial size (i.e. $64\times64$) of down-sampled ImageNet which is due to the fact that the feature size should maintain enough spatial size before each pooling layer. The fully connected layers size was changed to 2048 and the last pooling layer was eliminated. The results have shown in Figure \ref{fig:lfsr_benchmark} which shows that the proposed pruning method can preserve the accuracy even in high sparsity rates.

\begin{figure}[!b]
 \vspace{-0.1in}
\centering
  \includegraphics[scale=0.4]{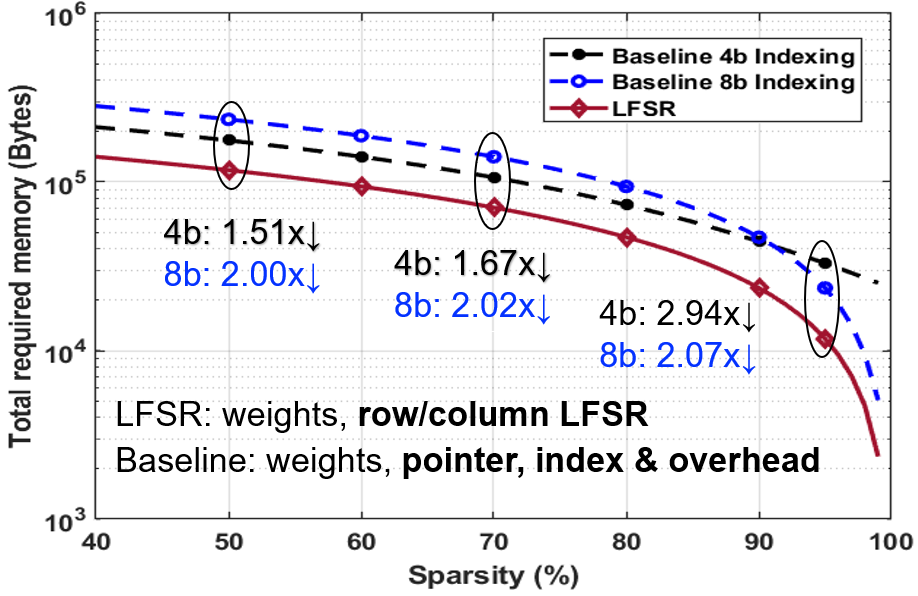}
    \vspace{-0.15in}
  \caption{Total required memory for our method and the proposed method with 4 and 8 bit precision at different levels of sparsity.}
  \label{fig:memory_usage}

\end{figure}
\vspace{-0.05in}
\subsection{65nm CMOS Hardware Implementation}

We have synthesized baseline and proposed methods with 65nm CMOS technology to measure hardware metrics. Implementation parameters are shown in Table \ref{tab:hardware_param}. The pre-layout analysis demonstrates 1.51$\times$ to 2.94$\times$ reduction in required memory footprint between proposed method and 4-8b indexed baseline pruning technique (Figure \ref{fig:memory_usage}). Besides memory, the overall system (memory, multiplier, accumulator and input/output buffers) parameters are also measured. The power and area measurements are demonstrated in Table \ref{tab:model_energy} and \ref{tab:model_area}. We observe a maximum of 63.96\% power savings and 68.18\% area savings across varying sparsity, indexing bit-widths and baseline designs. Although significant savings are observed for the proposed method, it should also be noted that the LFSR based column indexing introduces additional output buffer access (1 cycle read and 1 cycle write). This is included in our design and results. We note that additional power is negligible compared to the total power savings.

\begin{table}[t!]
\centering
\setlength\tabcolsep{3.9pt}
\caption{Measured Power ($mW$) of the overall system  (memory, multiplier, accumulator and  input/output  buffers)  for baseline and proposed method.}
\label{tab:model_energy}
\begin{tabular}{c|c|c|c|c|c|c|c}

 \multicolumn{2}{c|}{Network} & \multicolumn{2}{c|}{LeNet-300-100} & 
 \multicolumn{2}{c|}{LeNet-5}  & \multicolumn{2}{c}{modified VGG-16}\\
\cline{1-8}
\multicolumn{2}{c|}{\backslashbox{Sparsity}{Bit-width}} & 4bits & 8bits & 4bits & 8bits & 4bits& 8bits\\
\hline
\hline

 \multirow{3}{*}{\rotatebox[origin=c]{90}{\parbox[c]{1cm}{\centering Our Method}}} & 40\% & 439.9 & 439.9 & 102.9 &	102.9 &	37114 &	37114\\
\cline{2-8}

& 70\% & 118.2	&118.2 &	31.73 &	31.73 &	9294&	9294\\
\cline{2-8}
& 95\% & 46.74 &	46.74 &	15.91 &	15.91 &	3113 &	3113\\
\hline
\hline
\multirow{3}{*}{\rotatebox[origin=c]{90}{\parbox[c]{1cm}{\centering Baseline Method}}} & 40\% &643.4&	857.8 &	150.4 &	197.8	& 55639.1 &	74184\\
\cline{2-8}
&  70\% & 176.4 &	214.4 &	46.99 &	55.46 &	15239.4 &	18546\\
\cline{2-8}
& 95\% & 99.74 &	71.48&	30.11&	23.82&	8637.9&	6182\\
\hline
\hline
\multirow{3}{*}{\rotatebox[origin=c]{90}{\parbox[c]{1cm}{\centering Total Saving (\%)}}} & 40\% & \textbf{31.62} & \textbf{48.71}&	\textbf{31.56} &	\textbf{47.97} &	\textbf{33.29} &	\textbf{49.97}\\
\cline{2-8}
&  70\% & \textbf{32.98} &	\textbf{44.87}&	\textbf{32.47}&	\textbf{42.78}&	\textbf{39.01}&	\textbf{49.88}\\
\cline{2-8}
& 95\% & \textbf{53.13} &	\textbf{34.61} &	\textbf{47.15}&	\textbf{33.21}	&\textbf{63.96}&	\textbf{49.64}\\

\end{tabular}
\end{table}

\begin{table}[t!]
\centering
\setlength\tabcolsep{3.8pt}
\caption{Measured Area ($mm^2$) of the overall system  (memory, multiplier, accumulator
and  input/output  buffers)  for baseline and proposed method.}
\label{tab:model_area}
\begin{tabular}{c|c|c|c|c|c|c|c}

\multicolumn{2}{c|}{Network} &\multicolumn{2}{c|}{LeNet-300-100} & \multicolumn{2}{c|}{LeNet-5}  &\multicolumn{2}{c}{modified VGG-16} \\
\cline{1-8}
\multicolumn{2}{c|}{\backslashbox{Sparsity}{Bit-width}} & 4bits & 8bits & 4bits & 8bits & 4bits& 8bits
\\
\hline
\hline
 \multirow{3}{*}{\rotatebox[origin=c]{90}{\parbox[c]{1cm}{\centering Our Method}}} & 40\% & 1.69	&1.69	&0.37&	0.37	&146.12	&146.12\\
\cline{2-8}

& 70\% & 0.42	&0.42&	0.09	&0.09	&36.53&	36.53\\
\cline{2-8}
& 95\% &0.14	&0.14	&0.03	&0.03	&12.18	&12.18\\
\hline
\hline
\multirow{3}{*}{\rotatebox[origin=c]{90}{\parbox[c]{1cm}{\centering Baseline Method }}} & 40\% &2.55	&3.39	&0.57	&0.76	&219.20	&292.26\\
\cline{2-8}
&  70\% & 0.71	&0.86	&0.16	&0.20&	60.05&	73.07

\\
\cline{2-8}
& 95\% & 0.40	&0.29&	0.10&	0.07&	34.04&	24.36\\
\hline
\hline
\multirow{3}{*}{\rotatebox[origin=c]{90}{\parbox[c]{1cm}{\centering Total Saving (\%)}}} & 40\% & \textbf{33.62}&	\textbf{50.16} &	\textbf{34.62}	&\textbf{50.71} &	\textbf{33.34}&	\textbf{50.00}\\
\cline{2-8}
&  70\% &\textbf{40.15} &	\textbf{50.63}&	\textbf{43.15}&\textbf{	52.74}&	\textbf{39.16}&	\textbf{50.01}\\
\cline{2-8}
& 95\% & \textbf{65.16}&	\textbf{51.85}	&\textbf{68.18}&	\textbf{57.40}&	\textbf{64.23}&	\textbf{50.02}\\
\end{tabular}
\end{table}

\section{Conclusions}
In this paper we propose a new method of indexing to use a sparse network for inference, to enhance the memory usage and energy efficiency of DNNs. To achieve that, we have utilized an LFSR based indexing by generating two pseudo random sequence as indices instead of saving the indices in a separate memory. The generated indices are used to decide which weights need to be pruned and which ones to be retained. We show that our method can prune large networks without loss of accuracy. In addition, we demonstrated, a  maximum  of  63.96\%  power  savings  and  68.18\% area  savings  across  varying  sparsity,  indexing  bit-widths can be achieved.

\section{Acknowledgements}
This project was supported by the Semiconductor Research Corporation under grant JUMP CBRIC task ID 2777.004, 2777.005 and 2777.006.

\bibliographystyle{IEEEtran}
\bibliography{IEEEabrv,lfsrprun_iscas2020}


% George Kour and Raid Saabne.
% \newblock Real-time segmentation of on-line handwritten arabic script.
% \newblock In {\em Frontiers in Handwriting Recognition (ICFHR), 2014 14th
%   International Conference on}, pages 417--422. IEEE, 2014.

% George Kour and Raid Saabne.
% \newblock Fast classification of handwritten on-line arabic characters.
% \newblock In {\em Soft Computing and Pattern Recognition (SoCPaR), 2014 6th
%   International Conference of}, pages 312--318. IEEE, 2014.

% \bibitem{hadash2018estimate}
% Guy Hadash, Einat Kermany, Boaz Carmeli, Ofer Lavi, George Kour, and Alon
% \newblock Estimate and replace: A novel approach to integrating deep neural
%   networks with existing applications.
% \newblock {\em arXiv preprint arXiv:1804.09028}, 2018.

% \end{thebibliography}

\end{document}